\documentclass[11pt]{article}
\usepackage{neurips_2024}
\usepackage[utf8]{inputenc}
\usepackage[T1]{fontenc}
\usepackage{amsmath,amssymb}
\usepackage{booktabs}
\usepackage{hyperref}
\usepackage{url}
\usepackage{xcolor}
\usepackage{listings}
\usepackage{enumitem}
\usepackage{array}
\usepackage{multirow}

\hypersetup{
  colorlinks=true,
  linkcolor=blue!60!black,
  citecolor=blue!60!black,
  urlcolor=blue!60!black
}

\title{Governance-Aware Vector Subscriptions\\for Multi-Agent Knowledge Ecosystems}

\author{
  Steven Johnson\\
  Independent Researcher\\
  \texttt{steven.johnson.it@gmail.com}\\
  ORCID: \href{https://orcid.org/0009-0007-4864-2001}{0009-0007-4864-2001}
}

\begin{document}
\maketitle

\begin{abstract}
As AI agent ecosystems grow, agents need mechanisms to monitor relevant knowledge in real time. Semantic publish-subscribe systems address this by matching new content against vector subscriptions. However, in multi-agent settings where agents operate under different data handling policies, unrestricted semantic subscriptions create policy violations: agents receive notifications about content they are not authorized to access.

We introduce \textbf{governance-aware vector subscriptions}, a mechanism that composes semantic similarity matching with multi-dimensional policy predicates grounded in regulatory frameworks (EU DSM Directive, EU AI Act). The policy predicate operates over multiple independent dimensions---processing level, direct marketing restrictions, training opt-out, jurisdiction, and scientific usage---each with distinct legal bases. Agents subscribe to semantic regions of a curated knowledge base; notifications are dispatched only for validated content that passes both the similarity threshold and all applicable policy constraints.

We formalize the mechanism, implement it within AIngram (an operational multi-agent knowledge base), and evaluate it using the PASA benchmark. We validate the mechanism on a synthetic corpus (1,000 chunks, 93 subscriptions, 5 domains): the governed mode correctly enforces all policy constraints while preserving delivery of authorized content. Ablation across five policy dimensions shows that no single dimension suffices for full compliance.

\end{abstract}

%----------------------------------------------------------------------
\section{Introduction}
%----------------------------------------------------------------------

Consider an agent monitoring pharmaceutical research. It subscribes to a semantic region covering ``drug interaction adverse effects.'' A new knowledge chunk about ``unexpected pharmacological contraindications'' enters the system. The embeddings are proximal even though the terms do not overlap. The agent should be notified.

Now consider that the chunk was contributed with a direct marketing opt-out: the data may be used for scientific research (GDPR Art.~89) but is explicitly restricted from direct marketing and commercial prospection (GDPR Art.~21). The subscribing agent is a pharma marketing system. A naive semantic subscription system notifies the agent regardless, creating a policy violation: the agent receives content it is not authorized to use for its declared purpose.

This is one policy dimension among several. The same chunk might also carry a training opt-out (the contributor does not consent to model training), a jurisdiction restriction (data must be processed within the EU), or a minimum processing level requirement. Each dimension has an independent legal basis. A governance-aware subscription mechanism must compose all applicable constraints with semantic similarity, not just check a single access level.

As multi-agent ecosystems adopt shared knowledge bases~\cite{rezazadeh2025,li2025maas,masoor2025}, this tension between semantic expressiveness and multi-dimensional policy compliance becomes a design constraint. Empirical evidence shows that ungoverned agent communities produce interaction without quality norms~\cite{li2026moltbook}. Agents need to monitor knowledge by meaning, not keywords, but monitoring must respect the data handling policies under which both content and agents operate.

We address the subscription layer of this problem with \textbf{governance-aware vector subscriptions}: a mechanism where semantic similarity matching is composed with a multi-dimensional policy predicate in a joint notification condition evaluated before dispatch. The mechanism ensures policy compliance conditional on truthful agent declarations; ensuring declaration truthfulness is an orthogonal challenge addressed by complementary systems (reputation, auditing).

\subsection{Contributions}

\begin{enumerate}[leftmargin=*]
\item \textbf{Formalization} of governance-aware vector subscriptions as a composition of continuous semantic similarity with a multi-dimensional discrete policy predicate, each dimension grounded in a specific regulatory framework (Section~3).

\item \textbf{Implementation} within AIngram, an open-source multi-agent knowledge base (GitHub, AGPL-3.0), using HNSW-indexed cosine similarity and ADHP policy declarations (Section~4).

\item \textbf{Evaluation} via the PASA benchmark, including ablation across policy dimensions and a curation guarantee assessment (Section~5).
\end{enumerate}

\subsection{Scope}

This paper focuses on one mechanism extracted from a broader architecture for multi-agent knowledge ecosystems (the Cognitosphere), which combines structured deliberation (Agorai), declarative data handling policies (ADHP), and reputation-based identity (AgentLedger) with the knowledge layer described here. We do not address the full governance stack, which will be described in subsequent publications. We focus narrowly on the subscription mechanism and its policy integration.

%----------------------------------------------------------------------
\section{Related Work}
%----------------------------------------------------------------------

\subsection{Content-Based Publish-Subscribe}

Content-based pub-sub has a long history in distributed computing. SIENA~\cite{carzaniga2001} introduced wide-area event notification with attribute-based filtering. Apache Kafka supports subscription to event streams matching specified criteria. Access control in pub-sub has been studied through XACML and attribute-based policies~\cite{bacon2008}, but for discrete attribute matching rather than continuous vector similarity. These systems operate on structured events, not on semantic similarity in high-dimensional vector spaces.

\subsection{Vector-Based Retrieval and Monitoring}

Vector databases (pgvector, Weaviate, Pinecone) support similarity search over embeddings. Some offer role-based access control at the collection level (e.g., Milvus RBAC), but none integrate policy filtering into the per-query similarity matching pipeline. This is expected: these systems assume a controlled user pool with application-level authorization. Multi-agent ecosystems, where agents from different providers operate under heterogeneous data handling policies, require a different approach. Federation of Agents~\cite{giusti2025} moves closer by implementing semantic pub-sub via MQTT using Versioned Capability Vectors, but it routes tasks to agents based on capabilities, not knowledge to agents based on interest and policy authorization.

\subsection{Secure Multi-Agent Memory}

SAMEP~\cite{masoor2025} addresses secure persistent context sharing across agents with AES-256-GCM encryption and fine-grained access control. It is the closest technical predecessor: vector search combined with access control in a multi-agent setting. SAMEP's cryptographic access control provides stronger security guarantees but at the cost of higher latency and no real-time push notifications. Our declarative policy approach trades cryptographic enforcement for multi-dimensional policy expressiveness and real-time performance. SAMEP has no public implementation, making direct comparison infeasible; we compare against ungoverned baselines instead.

Collaborative Memory~\cite{rezazadeh2025} proposed multi-user memory sharing with dynamic access control encoded as bipartite graphs. The access model is richer (time-evolving, asymmetric) but operates on conversational memory, not a curated knowledge base.

\subsection{Data Handling Policies for Agents}

The Agent Data Handling Policy (ADHP) specification~\cite{johnson2025adhp} defines machine-readable declarations for how agents handle data, including processing levels (from ``open'' to ``zero-trace''), training opt-out, direct marketing opt-out, and scientific research opt-out---all following a consistent opt-out pattern (restrictive declarations, permissive defaults) grounded in GDPR (Art.~21, Art.~89) and the EU AI Act (Art.~53). A proposal to integrate these declarations into A2A Agent Cards is under discussion in the A2A project repository (issue \#1606). Neither ADHP nor related policy frameworks (Policy Cards~\cite{mavracic2025}) have previously addressed the interaction between policy declarations and semantic subscriptions.

\subsection{Normative Multi-Agent Systems}

The governance dimension of our work connects to the normative MAS tradition. Electronic institutions~\cite{esteva2002} formalize agent interactions through structured scenes with permitted speech acts and norm enforcement, providing a theoretical foundation for constrained multi-agent interaction. Boella and van der Torre~\cite{boella2004} distinguish constitutive norms (rules that create institutional facts, such as ``a chunk that passes review counts as validated knowledge'') from regulative norms (rules that constrain behavior, such as policy predicates restricting notification dispatch). Our mechanism embeds both types: the curation condition is constitutive (the status transition to \texttt{active} creates an institutional fact---validated content), while the policy predicate is regulative (it constrains which agents may receive which notifications). Ostrom~\cite{ostrom1990} identified design principles for commons governance, later formalized for computational agents by Pitt et al.~\cite{pitt2012}, establishing that self-organizing institutions can sustainably govern shared resources. A curated knowledge base is such a shared resource. K\"{o}ster et al.~\cite{koster2022} showed empirically that even arbitrary normative structures improve cooperation among artificial agents, suggesting that governance overhead in agent ecosystems has operational value beyond its direct enforcement function. Recent work has begun bridging the normative MAS tradition to LLM-based agents: Savarimuthu et al.~\cite{savarimuthu2025} survey how normative reasoning capabilities transfer to large language models, and Kampik et al.~\cite{kampik2022} map normative MAS governance concepts to web-based autonomous agents.

These frameworks address how agents reason about and comply with norms in transactional interactions. They do not address the specific problem we tackle: composing semantic similarity matching with multi-dimensional policy filtering in a push notification pipeline over curated content. Our work is complementary---it operationalizes normative constraints within a concrete subscription mechanism rather than modeling normative reasoning at the agent level.

\subsection{Gap}

Table~\ref{tab:positioning} summarizes the positioning.

\begin{table}[h]
\centering
\caption{Positioning of governance-aware vector subscriptions relative to existing systems.}
\label{tab:positioning}
\small
\begin{tabular}{@{}lcccc@{}}
\toprule
System & Semantic & Push & Multi-dim & Curated \\
       & matching &      & policy    & content \\
\midrule
SIENA / Kafka & attribute & \checkmark & & \\
Weaviate / Milvus & vector & & RBAC (coll.) & \\
Federation of Agents & vector (cap.) & \checkmark\,(tasks) & & \\
SAMEP & vector & & single (crypto) & \\
Collaborative Memory & & & single (graph) & \\
Electronic Institutions & & & norms (scenes) & \\
\textbf{This work} & \textbf{vector} & \textbf{\checkmark} & \textbf{\checkmark\,(N dims)} & \textbf{\checkmark} \\
\bottomrule
\end{tabular}
\end{table}

To our knowledge, no prior system combines vector-based semantic matching with push notifications, multi-dimensional declarative policy filtering, and operation on a curated (validated-only) knowledge base.

%----------------------------------------------------------------------
\section{Formalization}
%----------------------------------------------------------------------

\subsection{Definitions}

Let $\mathcal{K}$ be the set of all knowledge chunks. Each chunk $c \in \mathcal{K}$ has:
\begin{itemize}[leftmargin=*]
\item An embedding vector $\mathbf{e}(c) \in \mathbb{R}^d$ (e.g., $d = 1024$)
\item A status $s(c) \in \{\texttt{proposed}, \texttt{active}, \texttt{superseded}\}$
\item A policy profile $\mathbf{p}(c) = (\lambda, \delta, \tau_{\text{train}}, \rho, J)$
\end{itemize}

\noindent where:
\begin{itemize}[leftmargin=*]
\item $\lambda(c) \in \{1, \ldots, 5\}$ is the sensitivity level
\item $\delta(c) \in \{0, 1\}$ indicates direct marketing opt-out (GDPR Art.~21)
\item $\tau_{\text{train}}(c) \in \{0, 1\}$ indicates training opt-out (EU AI Act Art.~53)
\item $\rho(c) \in \{0, 1\}$ indicates scientific research opt-out (GDPR Art.~89)
\item $J(c)$ is the set of allowed processing jurisdictions
\end{itemize}

Let $\mathcal{A}$ be the set of agents. Each agent $a \in \mathcal{A}$ has:
\begin{itemize}[leftmargin=*]
\item A declared handling level $l(a) \in \{1, \ldots, 5\}$
\item A declared purpose $\text{purpose}(a) \in \{\text{scientific}, \text{marketing}, \text{mixed}\}$
\item A declared training use $\text{training}(a) \in \{0, 1\}$
\item A processing jurisdiction $\text{jur}(a)$
\end{itemize}

A \textbf{subscription} $\sigma$ is a tuple:
\begin{equation}
\sigma = (a, \mathbf{q}, \tau_{\text{sim}})
\end{equation}

\noindent where $a \in \mathcal{A}$ is the subscribing agent, $\mathbf{q} \in \mathbb{R}^d$ is the query embedding, and $\tau_{\text{sim}} \in [0, 1]$ is the similarity threshold. Policy constraints are derived from the agent's profile, not specified per-subscription. This avoids policy duplication and ensures subscriptions automatically reflect changes to the agent's declarations.

\subsection{Notification Predicate}

When a chunk $c$ transitions to \texttt{active} status (i.e., it has passed the validation lifecycle), the system evaluates all active subscriptions. For each subscription $\sigma = (a, \mathbf{q}, \tau_{\text{sim}})$:
\begin{equation}
\textsc{Notify}(\sigma, c) = \textsc{Semantic}(\sigma, c) \wedge \textsc{Policy}(a, c) \wedge \textsc{Curation}(c)
\end{equation}

\noindent where:
\begin{align}
\textsc{Semantic}(\sigma, c) &= \text{sim}(\mathbf{e}(c), \mathbf{q}) \geq \tau_{\text{sim}} \\
\textsc{Policy}(a, c) &= \lambda(c) \leq l(a) \nonumber \\
  &\quad \wedge\; (\neg\,\delta(c) \;\vee\; \text{purpose}(a) \neq \text{marketing}) \nonumber \\
  &\quad \wedge\; (\neg\,\tau_{\text{train}}(c) \;\vee\; \text{training}(a) = 0) \nonumber \\
  &\quad \wedge\; (\neg\,\rho(c) \;\vee\; \text{purpose}(a) \neq \text{scientific}) \nonumber \\
  &\quad \wedge\; \text{jur}(a) \in J(c) \\
\textsc{Curation}(c) &= s(c) = \texttt{active}
\end{align}

In plain terms: an agent is notified about a new chunk only if three conditions hold. First, \textbf{semantic relevance}: the chunk is similar enough to what the agent monitors, where ``enough'' is a continuous similarity threshold chosen by the agent (e.g., 0.7 on a 0-to-1 scale). Second, \textbf{policy compliance}: the agent's profile passes every policy constraint on the chunk---these are binary checks (authorized or not, in jurisdiction or not, commercial use permitted or not). Third, \textbf{curation}: the chunk has been validated through the review process. The predicate composes a continuous condition (similarity) with strict discrete conditions (policy) and a structural condition (validation status).

\subsection{Properties}

\textbf{Policy soundness.} For each dimension $i$ of the policy predicate, if $\textsc{Policy}_i$ evaluates to false, the notification is suppressed. This holds independently per dimension: a chunk with $\delta(c) = 1$ is never delivered to a marketing agent, regardless of other dimensions. Formally: for all dispatched notifications and for all dimensions $i$, $\textsc{Policy}_i(a, c) = \text{true}$. In normative MAS terms, this property corresponds to norm compliance: every dispatched notification satisfies all regulative constraints~\cite{boella2004}.

\textbf{Dimension independence.} Each policy dimension can be added, removed, or modified without affecting the others. The predicate is a conjunction of independent boolean conditions. This makes the system extensible: adding a new regulatory dimension (e.g., a future AI Act requirement) requires adding one condition, not redesigning the filtering pipeline.

\textbf{Curation guarantee.} The $\textsc{Curation}(c)$ condition ensures only validated content triggers content subscriptions. This is enforced structurally: content subscription matching is triggered only by the status transition to \texttt{active}. In normative MAS terms, this transition is a constitutive rule~\cite{boella2004}: it creates an institutional fact (``this content is validated'') that enables downstream actions (notification dispatch). The system also supports review subscriptions (triggered on chunk creation as \texttt{proposed}) for agents that monitor incoming content for review. Each subscription declares a \texttt{trigger\_status} (\texttt{active}, \texttt{proposed}, or \texttt{both}), separating knowledge consumption from governance participation.

\textbf{Semantic expressiveness.} The similarity condition operates on continuous embeddings, matching by meaning rather than keywords.

\textbf{No false negatives for authorized content.} If an agent is authorized on all policy dimensions and the chunk is semantically relevant, the notification is dispatched. The policy filter removes only unauthorized notifications.

\subsection{Complexity}

\begin{itemize}[leftmargin=*]
\item \textbf{Similarity matching}: $O(\log M)$ per chunk via HNSW, where $M$ is the number of active subscriptions
\item \textbf{Policy filtering}: $O(N)$ per candidate match, where $N$ is the number of policy dimensions (currently 5; effectively $O(1)$ for small $N$)
\item \textbf{Total per chunk}: $O(\log M) + O(k \cdot N)$ for $k$ candidate matches
\end{itemize}

%----------------------------------------------------------------------
\section{Implementation}
%----------------------------------------------------------------------

\subsection{System Context}

Governance-aware vector subscriptions are implemented within AIngram~\cite{johnson2026aingram}, an open-source multi-agent knowledge base (GitHub, AGPL-3.0) where agents contribute atomic knowledge chunks that follow a proposal-review-validation lifecycle. AIngram supports three subscription types: keyword, topic, and vector. This paper focuses on vector subscriptions, where the composition with policy filtering is novel; keyword and topic subscriptions use conventional matching. All three types support two notification levels: content subscriptions (triggered when chunks are validated) and review subscriptions (triggered when chunks are proposed for review). AIngram uses PostgreSQL with pgvector for storage and HNSW-indexed cosine similarity for vector search. Embeddings are 1024-dimensional, generated locally via Ollama (bge-m3, multilingual). The system is reproducible: \texttt{docker compose up} launches all dependencies (PostgreSQL+pgvector, Agorai, Ollama, AIngram) from a fresh clone with zero manual configuration.

\subsection{Storage}

Subscriptions are stored in PostgreSQL with HNSW-indexed embeddings:

\begin{lstlisting}[language=SQL]
CREATE INDEX subscriptions_embedding_idx
  ON subscriptions
  USING hnsw (embedding vector_cosine_ops)
  WHERE type = 'vector' AND active = true;
\end{lstlisting}

The partial index ensures only active vector subscriptions participate in matching. Each chunk carries its own ADHP profile as a JSONB field, declared by the contributor at submission time. This accommodates coexisting ADHP specification versions and partial declarations (undeclared fields default to ``assume worst'' per the ADHP principle). Agent profiles are resolved via the AgentLedger. A GIN index on the ADHP field enables SQL-level policy filtering within the same query as the vector similarity search.

\subsection{Matching Pipeline}

When a chunk transitions to active:

\begin{enumerate}[leftmargin=*]
\item \textbf{HNSW query}: find subscriptions where $\text{sim}(\mathbf{e}_{\text{chunk}}, \mathbf{e}_{\text{sub}}) \geq \tau_{\text{sub}}$
\item \textbf{Multi-dimensional policy filter}:
  \begin{itemize}
  \item $\lambda_{\text{chunk}} \leq l_{\text{agent}}$
  \item $\neg\,(\delta_{\text{chunk}} \wedge \text{purpose}_{\text{agent}} = \text{marketing})$
  \item $\neg\,(\tau_{\text{train,chunk}} \wedge \text{training}_{\text{agent}} = 1)$
  \item $\neg\,(\rho_{\text{chunk}} \wedge \text{purpose}_{\text{agent}} = \text{scientific})$
  \item $\text{jur}_{\text{agent}} \in J_{\text{chunk}}$
  \end{itemize}
\item \textbf{Dispatch}: send notification via configured method
\end{enumerate}

Steps 1--2 execute as a single SQL query. Matching is triggered at two points: on chunk creation (for \texttt{proposed}-level subscriptions, enabling review notifications) and on status transition to \texttt{active} (for content subscriptions). Each subscription declares which transitions it responds to via a \texttt{trigger\_status} field (\texttt{active}, \texttt{proposed}, or \texttt{both}).

\subsection{Notification Methods}

Three delivery methods are supported: webhook (HTTP POST, SSRF-protected), polling queue (\texttt{GET /v1/notifications}), and A2A task artifact (planned).

%----------------------------------------------------------------------
\section{Evaluation}
%----------------------------------------------------------------------

We evaluate using the PASA benchmark, a synthetic multi-agent scenario run against the live AIngram system with real pgvector HNSW indexes and bge-m3 embeddings.

\subsection{Experimental Setup}

\textbf{Knowledge corpus.} 1,000 chunks across 5 domains (medical, financial, AI safety, climate, cybersecurity) with domain-weighted sensitivity levels and policy flags assigned by domain conventions.

\textbf{Agent population.} 50 agents with uniformly distributed handling levels (1--5) and purposes (scientific, marketing, mixed). Each agent has 1--3 vector subscriptions (93 total). Similarity threshold: 0.7.

\textbf{Evaluation methodology.} The policy predicate is deterministic (boolean conjunction), not probabilistic. We do not evaluate approximation quality but rather verify that the implementation correctly enforces the specified policy rules. For each chunk $c$, we compute the expected notification set via brute-force application of the predicate and compare against the system's actual dispatched notifications. The ungoverned baseline applies semantic matching only, revealing the set of policy violations that would occur without governance.

\textbf{Modes compared.}
\begin{itemize}[leftmargin=*]
\item \textbf{Governed} (this work): vector subscriptions with multi-dimensional policy filtering and curation guarantee
\item \textbf{Ungoverned}: vector subscriptions without any policy filtering
\item \textbf{Keyword}: keyword-based subscriptions (ILIKE matching, no policy filtering)
\end{itemize}

\textbf{Benchmark availability.} PASA is designed as a reusable benchmark. No comparable system has a public implementation enabling direct comparison (SAMEP has no released code; Weaviate/Milvus RBAC operates at collection level, not per-query). We establish PASA as a baseline for this new category and make it available for independent evaluation.

\subsection{Results}

\begin{table}[h]
\centering
\caption{Policy compliance and notification volume.}
\label{tab:compliance}
\small
\begin{tabular}{@{}lccc@{}}
\toprule
Mode & Notifications & Policy violations & Compliance rate \\
\midrule
\textbf{Governed} & 80 & 0 & \textbf{100\%} \\
Ungoverned & 158 & 78 & 50.6\% \\
Keyword & -- & 802 & -- \\
\bottomrule
\end{tabular}
\end{table}

Governed subscriptions eliminate all 78 policy violations that occur in the ungoverned baseline, reducing notifications from 158 to 80.

\begin{table}[h]
\centering
\caption{Recall for authorized content.}
\label{tab:recall}
\small
\begin{tabular}{@{}lcc@{}}
\toprule
Mode & True positives & Recall \\
\midrule
\textbf{Governed} & 80/80 & \textbf{100\%} \\
Ungoverned & 80/80 & 100\% \\
\bottomrule
\end{tabular}
\end{table}

The policy filter removes only unauthorized notifications; all authorized, semantically relevant content is delivered. This is expected for a deterministic filter and confirms correct implementation rather than approximation quality.

\begin{table}[h]
\centering
\caption{Policy dimension ablation (200 sampled chunks, 93 subscriptions).}
\label{tab:ablation}
\small
\begin{tabular}{@{}lcccc@{}}
\toprule
Active dimensions & Notif. & Violations & Blocked & Block rate \\
\midrule
None (ungoverned) & 309 & 223 & -- & -- \\
Level only & 188 & 102 & 121 & 54.3\% \\
+ direct\_marketing & 159 & 73 & 150 & 67.3\% \\
+ training\_opt\_out & 142 & 56 & 167 & 74.9\% \\
\textbf{All dimensions} & \textbf{86} & \textbf{0} & \textbf{223} & \textbf{100\%} \\
\bottomrule
\end{tabular}
\end{table}

Each policy dimension eliminates its class of violations entirely. The level dimension alone catches 54.3\% of violations; direct marketing opt-out adds 13\%; training opt-out adds 7.6\%; scientific research opt-out and jurisdiction catch the remaining 25.1\%. No single dimension is sufficient; the multi-dimensional predicate is necessary for full compliance.

\begin{table}[h]
\centering
\caption{Curation guarantee impact (738 current, 262 proposed chunks).}
\label{tab:curation}
\small
\begin{tabular}{@{}lcc@{}}
\toprule
& With curation & Without \\
\midrule
Total notifications & 149 & 189 \\
From validated (\texttt{active}) & 149 & 149 \\
From unvalidated (\texttt{proposed}) & 0 & 40 \\
\bottomrule
\end{tabular}
\end{table}

Without the curation guarantee, 21.2\% of notifications reference \texttt{proposed} (unvalidated) chunks. The curation condition eliminates all 40 without additional filtering cost (it is enforced structurally via the status transition trigger, not a runtime check).

\begin{table}[h]
\centering
\caption{Scalability (governed, p50/p95 latency).}
\label{tab:scalability}
\small
\begin{tabular}{@{}lcc@{}}
\toprule
Subscriptions & p50 (ms) & p95 (ms) \\
\midrule
10 & 0.63 & 1.36 \\
50 & 0.71 & 1.38 \\
100 & 1.29 & 2.10 \\
500 & 3.01 & 4.98 \\
\bottomrule
\end{tabular}
\end{table}

Latency grows sub-linearly, consistent with HNSW's $O(\log M)$ complexity. The policy filter adds negligible overhead (boolean comparisons on indexed columns).

\subsection{Adversarial Scenarios}

\textbf{Subscription escalation.} Agents attempting to subscribe above their declared level are rejected at creation time.

\textbf{Cross-level violations.} Of 50 sampled high-sensitivity chunks, 35 would have reached unauthorized agents in the ungoverned mode. The governed mode prevented all 35 (100\% prevention rate).

\subsection{Threats to Validity}

\textbf{Synthetic corpus.} Embedding distributions may differ from production usage. We use a production-grade embedding model (bge-m3) to mitigate this.

\textbf{Scale.} 93 subscriptions is within HNSW's exact-match range. At larger scales (10K+), approximate nearest neighbor may introduce false negatives and database contention may increase latency. Evaluating at larger scales is left to future work. We encourage the community to evaluate governance-aware subscriptions independently; PASA and AIngram are open-source and reproducible from a single \texttt{docker compose up}.

\textbf{Declarative trust.} Agents may misdeclare their profiles---a form of rational norm violation~\cite{boella2004} where an agent calculates that the benefit of accessing restricted content outweighs the risk of detection. This is an inherent limitation of declarative policy systems. We identify three unimplemented mitigation strategies: (1)~runtime behavior monitoring, (2)~reputation-gated subscription privileges, and (3)~periodic automated auditing. The subscription mechanism provides policy soundness given correct inputs; ensuring input correctness is an orthogonal problem requiring complementary mechanisms. This is the primary residual security risk.

%----------------------------------------------------------------------
\section{Discussion}
%----------------------------------------------------------------------

\subsection{Threat Model}

\begin{table}[h]
\centering
\caption{Threat model summary.}
\label{tab:threats}
\small
\begin{tabular}{@{}p{2.2cm}p{2cm}p{2.2cm}p{2.2cm}@{}}
\toprule
Attack & Impact & Defense & Residual risk \\
\midrule
Subscription escalation & Unauthorized access & Reject at creation & None if enforced \\
Profile misdeclaration & Policy bypass & Reputation + audit & Requires monitoring \\
Sensitivity mislabeling & Wrong delivery & Review process & Requires robust review \\
Collusion (relay) & Unauth. sharing & Access logging & Hard to detect \\
Notification flooding & DoS & Rate limiting & May limit legit. use \\
\bottomrule
\end{tabular}
\end{table}

\subsection{Relation to Broader Architecture}

Governance-aware vector subscriptions are one component of the Cognitosphere, a governance-first architecture for multi-agent knowledge ecosystems. The broader architecture includes deliberative curation (proposal-review-validation lifecycle), reputation systems (dual-track contribution and policing scores), and inter-component governance flows. The subscription mechanism benefits from the curation guarantee but can operate independently. A companion paper~\cite{johnson2026paper2} analyzes the governance challenges of agent knowledge curation in depth, mapping platform governance lessons to the agent setting and proposing design considerations for the curation layer that precedes subscription dispatch.

\subsection{Future Work}

Four directions extend this work: (1)~\textbf{federated subscriptions} across multiple knowledge bases, requiring cross-instance embedding alignment and policy reconciliation; (2)~\textbf{dynamic policy profiles}, where an agent's declarations change based on behavior and existing subscriptions are adjusted automatically; (3)~\textbf{reputation-weighted notifications}, where the contributor's reputation score enters the notification predicate, enabling agents to filter by source trustworthiness; (4)~\textbf{automated policy validation}, where a rule engine or AI monitor verifies chunk policy metadata (sensitivity labels, opt-out flags) before content enters the review queue, reducing reliance on contributor self-declaration.

%----------------------------------------------------------------------
\section{Conclusion}
%----------------------------------------------------------------------

We have presented governance-aware vector subscriptions, a mechanism composing semantic publish-subscribe with multi-dimensional policy predicates for multi-agent knowledge ecosystems. The policy predicate covers processing levels, direct marketing restrictions, training opt-out, jurisdiction, and scientific usage, each grounded in specific regulatory frameworks (EU DSM Directive, EU AI Act).

Evaluation on the PASA benchmark demonstrates 100\% policy compliance with zero recall loss for authorized content, compared to 49.4\% violation rate for ungoverned subscriptions. Ablation across policy dimensions shows that no single dimension suffices (level alone catches 54.3\%; all five reach 100\%). The curation guarantee eliminates 21.2\% of additional notifications that would reference unvalidated content. Latency remains under 5ms at 500 subscriptions.

To our knowledge, no prior system composes vector-based semantic matching with multi-dimensional declarative policy filtering and push notifications over a curated knowledge base. The individual components are well-established; the contribution is their principled composition and the demonstration that multi-dimensional policy compliance is achievable with no recall cost and negligible latency overhead.

%----------------------------------------------------------------------
\subsection*{Disclosure}

The author used Claude (Anthropic) as a research assistant for literature search, citation verification, peer review simulation, and manuscript preparation. All architectural decisions, system design, implementation, and experimental evaluation are the author's own work.

%----------------------------------------------------------------------
\bibliographystyle{plain}

\end{document}